\documentclass[numbers, sort]{ecai}
\usepackage{times}
\usepackage{graphicx}
\usepackage{latexsym}

\usepackage{graphicx}
\usepackage{amsmath}
\usepackage{amsthm}
\usepackage{caption}
\usepackage{algorithm}
\usepackage{algorithmic}
\usepackage{color}
\usepackage{tabu}
\usepackage{multirow,multicol,ctable}

\usepackage[utf8]{inputenc}
\usepackage{times}
\usepackage{helvet}
\usepackage{courier}
\usepackage{url}
\usepackage{amsmath}
\usepackage{amsthm}
\usepackage{amssymb}
\usepackage{graphicx}
\usepackage[export]{adjustbox}
\usepackage{multicol}
\usepackage{braket}
\usepackage{array}
\usepackage{tabu}
\usepackage{setspace}
\usepackage{verbatim}

\begin{document}

\title{RPN: A Residual Pooling Network for Efficient Federated Learning}

\author{Anbu Huang\institute{ Webank AI Lab, China, email: stevenhuang@webank.com} \and 
Yuanyuan Chen\institute{ Nanyang Technological University, Singapore} \and
Yang Liu\institute{ Webank AI Lab, China, email: yangliu@webank.com} \and 
Tianjian Chen\institute{ Webank AI Lab,
China, email: tobychen@webank.com} \and 
Qiang Yang\institute{ The Hong Kong University of Science and Technology,
Hong Kong, email: qyang@cse.ust.hk} 
}

\maketitle
\bibliographystyle{ecai}

\begin{abstract}
Federated learning is a distributed machine learning framework which enables different parties to collaboratively train a model while protecting data privacy and security. Due to model complexity, network unreliability and connection in-stability, communication cost has became a major bottleneck for applying federated learning to real-world applications. Current existing strategies are either need to manual setting for hyperparameters, or break up the original process into multiple steps, which make it hard to realize end-to-end implementation. In this paper, we propose a novel compression strategy called Residual Pooling Network (RPN). Our experiments show that RPN not only reduce data transmission effectively, but also achieve almost the same performance as compared to standard federated learning. Our new approach performs as an end-to-end procedure, which should be readily applied to all CNN-based model training scenarios for improvement of communication efficiency, and hence make it easy to deploy in real-world application without much human intervention.
\end{abstract}

\section{INTRODUCTION}
In the past decade, Deep Convolutional Neural Networks (DCNN) have shown powerful representation and learning capabilities \cite{Goodfellow:2016:DL:3086952,0483bd9444a348c8b59d54a190839ec9}, and achieve unprecedented success in many commercial applications, such as computer vision \cite{Krizhevsky:2012:ICD:2999134.2999257,DBLP:journals/corr/HeZRS15}, nature language processing \cite{mikolov2013efficient,DBLP:journals/corr/abs-1810-04805,DBLP:journals/corr/BojanowskiGJM16}, speech recognition \cite{DBLP:journals/corr/HannunCCCDEPSSCN14,DBLP:journals/corr/ZhangCJ16,DBLP:journals/corr/OordDZSVGKSK16}, etc. Same as traditional machine learning algorithms, the success of deep neural network is partially driven by big data availability. However, in the real world, with the exception of few industries, most fields have only limited data or poor quality data. What's worse, due to industry competition, privacy security, and complicated administrative procedures, It is almost impossible to train centralized machine learning models by integrating the data scattered around the countries and institutions \cite{DBLP:journals/corr/abs-1902-04885}. 

At the same time, with the increasing awareness of data privacy, the emphasis on data privacy and security has became a worldwide major issue. News about leaks on public data are causing great concerns in public media and governments. In response, countries across the world are strengthening laws in protection of data security and privacy. An example is the General Data Protection Regulation (GDPR) \cite{Voigt:2017:EGD:3152676} enforced by the European Union on May 25, 2018. Similar acts of privacy and security are being enacted in the US and China. 

To decouple the need for model training from the need for storing large data in the central database, a new machine learning framework called federated learning was proposed \cite{DBLP:journals/corr/McMahanMRA16}. Federated learning provides a promising approach for model training without compromising data privacy and security \cite{DBLP:journals/corr/abs-1902-01046,DBLP:journals/corr/McMahanMRA16,DBLP:journals/corr/KonecnyMYRSB16}. Unlike traditional centralized training procedures, federated learning requires each client collaboratively learn a shared model using the training data on the device and keeping the data locally, under federated learning scenario, model parameters (or gradients) are transmitted between the server side and the clients on each round. Due to the frequent exchange of data between the central server and the clients, coupled with network unreliability and connection instability, communication costs have became the main constraints and limitations of federated learning performance.

In this paper, we propose a new compression strategy called Residual Pooling Network (RPN) to address the communication costs problem by parameter selection and approximation. As we will see below, our approach can significantly reduce the quantity of data transmission on one hand, and still able to maintain high-level model performance on the other hand, more importantly, our approach can make compression process as an end-to-end procedure, which make it easy to deploy in real-world application without human intervention.

\textbf{\textit{Contributions:}} Our main contributions in this paper are as follows:
\begin{itemize}
\item we propose a practical and promising approach to improve communication efficiency under federated learning setting. Our approach not only reduces the amount of data being uploaded, but also reduces the amount of data being download.

\item we propose a general solution for CNN-based model compression under federated learning framework.

\item Unlike parameter-encryption based approach, our algorithm based on parameter approximation and parameter selection, which can keep data security without compromising communication efficiency, and is easy to deploy into large-scale systems.

\end{itemize}

The rest of this paper is organized as follows: We first review the related works of federated learning and current communication efficiency strategies. Then, we introduce our approach in detail. After that, we present the experimental results, followed by the conclusion and a discussion of future work.

\section{Related Work}

In this section, we will introduce some backgrounds and related works of our algorithm, including federated learning and communication costs. 

\subsection{Federated Learning}
\label{section:fl}

\begin{figure}[htb]
    \centering
    \includegraphics[width=3in]{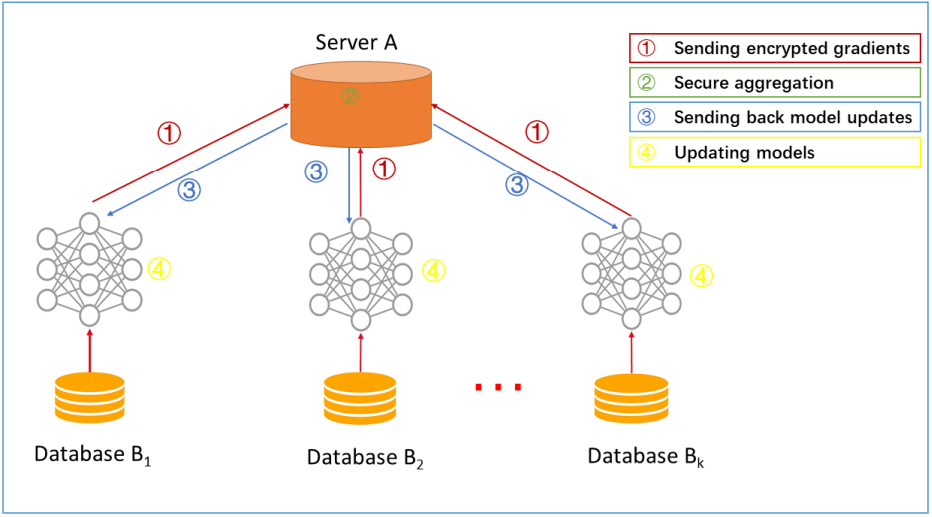}
    \caption{Federated Learning Architecture \cite{DBLP:journals/corr/abs-1902-04885}}
    \label{fig:fl_framework}
\end{figure}

In the traditional machine learning approach, data collected by different clients (IoT devices, smartphones, etc.) is uploaded and processed centrally in a cloud-based server or data center. However, due to data privacy and data security, sending raw data to the central database is regarded as unsafe, and violate the General Data Protection Regulation (GDPR) \cite{Voigt:2017:EGD:3152676}. To decouple the need for machine learning from the need for storing large data in the central database, a new machine learning framework called federated learning was proposed, a typical federated learning framework is as shown in Figure \ref{fig:fl_framework}. 

In federated learning scenario, each client update their local model based on local datasets, and then send the updated model’s parameters to the server side for aggregation, these steps are repeated in multiple rounds until the learning process converges. since local datasets remain on personal devices during training process, data security can be guaranteed.

In general, the standard federated learning training process including the following three steps \cite{DBLP:journals/corr/abs-1902-04885,lim2019federated}:

\begin{itemize}
\item \textbf{\textit{Local Model Training}}: Denote $t$ as the current iteration round. $N$ is the number of clients, $G_i^t$ is the initial model of this round for client $i$, $G_i^t$ parameterized by $w_i^t$, that is to say $G_i^t=f(w_i^t)$, $D_i$ is the local datasets of client $i$. Based on $D_i$, each client respectively update the local model parameters from $w_i^t$ to $w_i^{t+1}$, the updated local model parameters $w_i^{t+1}$ are then subsequently sent to the server side. 

To improve communication efficiency, when each round begins, we usually don't require all clients execute local model training, only a subset of $M$ clients should be selected, \cite{DBLP:journals/corr/McMahanMRA16} pointed out that such strategy does not reduce model performance.

\item \textbf{\textit{Global Aggregation}}: The server side aggregates the local model parameters of selected clients, and calculate the global model, we can formalize our aggregation algorithm as follows:

\begin{equation}\label{average}
    w^{t+1}=\sum_{i=1}^{M}{f(w_i^{t+1})}
\end{equation}

Where $f$ is aggregation operator, some sensible choices of aggregation operator including FedAvg \cite{DBLP:journals/corr/McMahanMRA16}, Byzantine Gradient Descent \cite{DBLP:journals/corr/ChenSX17}, Secure Aggregation \cite{Bonawitz:2017:PSA:3133956.3133982}, Automated Update of Weights.

\item \textbf{\textit{Update Local Model}}: When the aggregation is completed, the server side select a subset of clients again, and send global model $G^{t+1}$ back to the selected clients for next iteration and repeat this cycle until converge.

\end{itemize}

\subsection{Communication Costs}

In the standard federated learning training procedure, all model parameters should be exchanged between the server side and the clients, for complicated deep learning model, such as CNN, may comprise millions of parameters, coupled with network unreliability and connection in-stability, making the communication cost a significant bottleneck, as such, how to improve the communication efficiency of federated learning has became an urgent task. The total number of bits that have to be transmitted during model training is given by:

\begin{equation}\label{sum}
    S_{total}=\sum_{t=1}^{T}{\{F(G^{t})+\sum_{i=1}^{M}{F(G^{t}_{i})}\}}
\end{equation}

where $T$ is the total number of iterations between the server size and the clients, $M$ is the number of clients selected by the server side to update at round $t$, $G^t$ denotes global model after $t$ times aggregation, $F(G^{t})$ is the selective parameter bits download to client side, similarly, $F(G^{t}_{i})$ is the selected parameter bits of client $i$ used to upload to server side. Using equation \ref{sum} as a reference, we can classify the current research topics on communication efficiency by the following four aspects:

\bigskip
\textbf{\textit{Iterations frequency:}}  One feasible solution to improve communication efficiency is to reduce the number of communications (see $N$ in equation \ref{sum}) between the server side and the clients. McMahan et al.\cite{DBLP:journals/corr/SureshYMK16} proposed an iterative model averaging algorithm called Federated Averaging (FedAvg), and points out that each client can iterate the local SGD update multiple times before the averaging step, thus reducing the total number of communication rounds. The experiments showed a reduction in required communication rounds by $10$ to $100\times$ as compared to FedSGD.

\bigskip
\textbf{\textit{Pruning:}} Another research topic is to use model compression algorithm to reduce data transmission (see $F(G^{t})$ and $F(G^{t}_{i})$ in equation \ref{sum}), This is a technique commonly used in distributed learning \cite{lim2019federated,wang2018atomo}. A naive implementation requires that each client sends full model parameters back to the server in each round, and vice versa. Inspired by deep learning model compression algorithms\cite{DBLP:journals/corr/HanMD15}, distributed model compression approaches have been widely studied. Strom el.al\cite{strom2015scalable,tsuzuku2018variance} proposed an approach in which only gradients with magnitude greater than a given threshold are sent to the server, however, it is difficult to define threshold due to different configurations for different tasks. In a follow-up work, Aji et al.\cite{aji2017sparse} fixed transmission ratio p, and only communicate the fraction p entries with the biggest magnitude of each gradient. 

Konecny et.al\cite{DBLP:journals/corr/KonecnyMYRSB16} proposed low rank and subsampling of parameter matrix to reduce data transmission. Caldas et.al \cite{DBLP:journals/corr/abs-1812-07210} extend on the studies in \cite{DBLP:journals/corr/KonecnyMYRSB16} by
proposing lossy compression and federated dropout to reduce server-to-participant communication costs. 

\bigskip
\textbf{\textit{Importance-based Update:}} This strategy involves selective communication such that only the important or relevant updates are transmitted in each communication round. The authors in \cite{216799} propose the edge Stochastic Gradient Descent (eSGD) algorithm that selects only a small fraction
of important gradients to be communicated to the FL server
for parameter update during each communication round. the authors in \cite{CMFL2019} propose the
Communication-Mitigated Federated Learning (CMFL) algorithm that uploads only relevant local updates to reduce
communication costs while guaranteeing global convergence.

\bigskip
\textbf{\textit{Quantization:}} quantization is also a very important compression technique. Konecny et.al\cite{DBLP:journals/corr/KonecnyMYRSB16} proposed probabilistic quantization, which reduces each scalar to one bit (maximum value and minimum value). Bernstein et al.\cite{DBLP:journals/corr/abs-1802-04434,DBLP:journals/corr/abs-1810-05291} proposed signSGD, which quantizes gradient update to its corresponding binary sign, thus reducing the size by a factor of $32\times$. Sattler et.al \cite{DBLP:journals/corr/abs-1903-02891} proposed a new compression technique, called Sparse Ternary Compression (STC), which is suitable for non-iid condition. 

\bigskip
Aforementioned approaches are feasible strategies to improve communication efficiency during federated learning training procedure, some of which can reduce the amount of data transferred, while some other can reduce the number of iterations. However, all these solutions are either need to manual setting for hyper-parameter, or break up the original process into multiple steps, as such, cannot implement end-to-end workflow.

\section{Methodology}
\begin{figure*}[htb]
    \centering
    \includegraphics[width=5.5in]{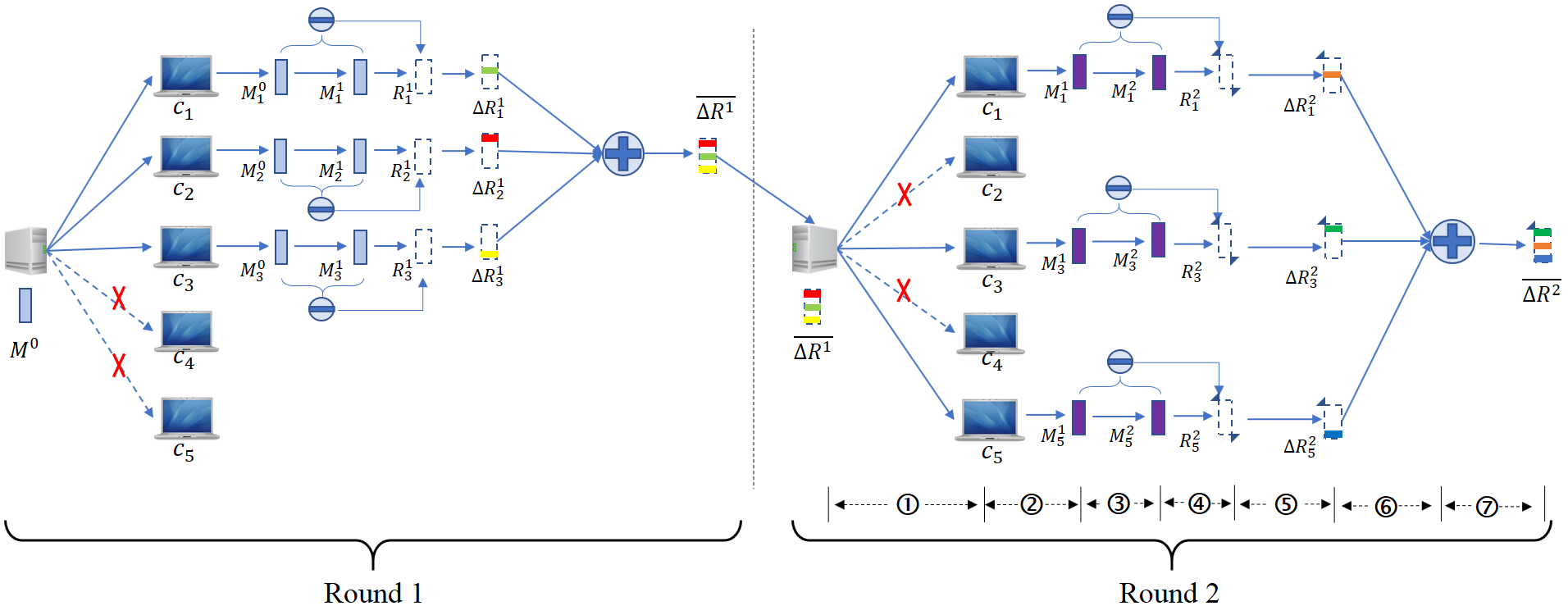}
    \caption{New federated learning workflow of our approach: (1) select clients for local model update. (2) recover local model. (3) local model training based on local datasets. (4) calculate residual network. (5) spatial pooling. (6) send rpn to server side, and do aggregation. (7) send rpn back to selected clients, and repeat this cycle.}
    \label{fig:new_workflow}
\end{figure*}

In this section, we formalize our problem definition, and show how to use Residual Pooling Network (RPN) to improve communication efficiency under federated learning framework, our new federated learning framework is shown in Figure \ref{fig:new_workflow}.

\subsection{Symbol Definition}

For the sake of consistence in this paper, we will reuse symbol definitions of section \ref{section:fl} in the following discussion. Suppose that our federated learning system consists of $N$ clients and one central server $S$, as illustrated in Figure \ref{fig:fl_framework}, $C_i$ denotes the $i$th client, each of which has its own datasets, indicated by $D_i=\{X_{i}, Y_{i}\}$ respectively, ${G}^{t}$ represents global model after $t$ times aggregation, ${G}^{t}$ is parameterized by ${w}^{t}$, that is to say: 

\begin{equation}\label{rep}
    {G}^{t}=f({w}^{t})  
\end{equation}

For simplicity, and without loss of generality, supposed we have completed $(t-1)$ times aggregation, and currently training on round $t$, client $C_i$ need to update local model from ${G_i}^{t-1}$ to ${G_i}^{t}$ based on local dataset $D_i$. The local model objective function is as follows: 
\begin{equation}\label{loss}
    \min_{w \in R^d}{l(Y_i, f(X_i;{w}))}
\end{equation}

The goal of client $i$ in iteration $t$ is to find optimal parameters $w$ that minimize the loss function of equation \ref{loss}.

let $R_i^t$ represents the residual network, which is given by: 

\begin{equation}\label{residual}
    R^t_i = G^{t}_i - G^{t-1}_i
\end{equation}

Execute spatial mean pooling on $R^t_i$, it would be changed to $\Delta{R^t_i}$. 




%

\subsection{Residual Pooling Network}

The standard federated learning requires exchange all parameters between the server side and the clients, but actually, many literatures \cite{strom2015scalable,tsuzuku2018variance,DBLP:journals/corr/HanMD15} had shown that not all parameters are sensitive to model updates in distributed model training. Under this assumption, we propose Residual pooling network, which compress data transmission through parameter approximation and parameter selection, RPN consists of the following two components:

\begin{itemize}
\item \textbf{Residual Network:} The difference of model parameters before and after the model update is called Residual network. Residual network is used to Capture and evaluate the contribution of different filters, the sum of kernel elements is regarded as the contribution of this filter, only magnitude greater than a given threshold are sent to the server side.

The purpose of Residual Network is to capture the changes in parameters during training, the author in \cite{DBLP:journals/corr/HeZR014} shown that different filters would have different responses to different image semantics. In light of different distribution of federated clients, each client $i$ would capture different semantic of local dataset $D_i$, as such, the changes of each layer in local model $G^t_i$ are different, which makes it no need to upload all model parameters.

\begin{figure}[h]
    \centering
    \includegraphics[width=3.3in]{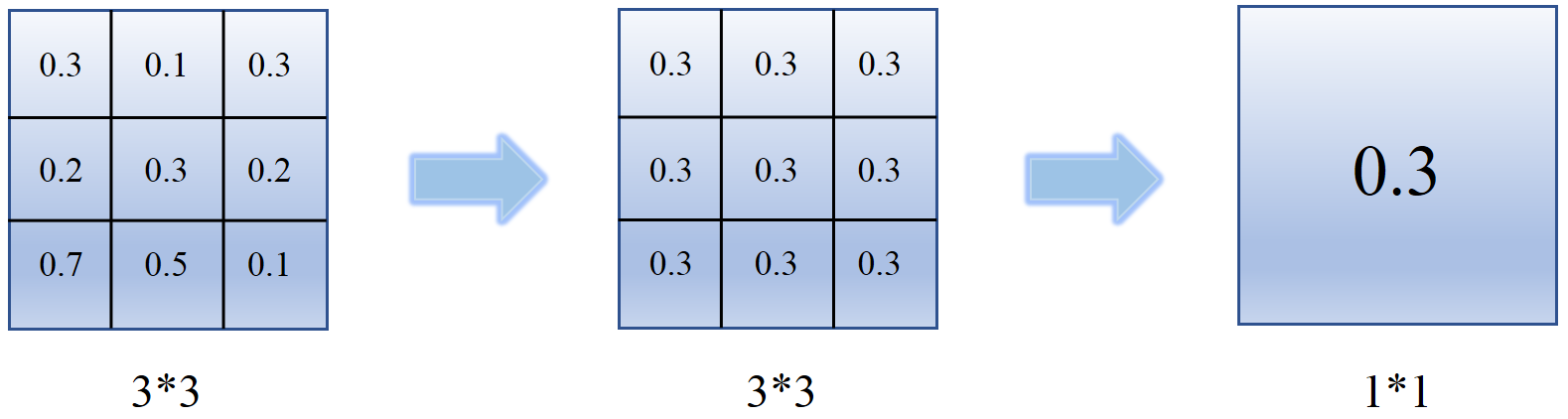}
    \caption{Spatial mean pooling, we calculate the mean value of the kernel, use this value as new kernel with size (1, 1)}
    \label{fig:pool}
\end{figure}

\item \textbf{Spatial pooling:} Spatial pooling is used to further compress model parameters. After we select high sensitive filters through residual network, using spatial mean pooling to compress each filter size from $(size, size)$ to $(1, 1)$, as shown in Figure \ref{fig:pool}.

Accord to \cite{DBLP:journals/corr/IoffeS15}, as the parameters of the previous layers change, the distribution of each layer’s inputs changes during model training, we refer to this phenomenon as internal covariate shift, the feasible solution to address the problem is using batch normalization to normalize each layer inputs. Similarly, during federated learning scenario, after local model training, the change in the distribution of network activations due to the change in network parameters during training, spatial pooling of convolutional layer make it normalize each layer by average operation, which makes it reduce the internal covariate shift to some extent.
\end{itemize}


\subsection{Residual Pooling Network - Server perspective}

\begin{algorithm}
   \caption{RPN: FL\_Server}
   \label{algo_server}
   \setstretch{1.3} 
\begin{algorithmic}
   \STATE Initialize $G^0$, and set $\overline{\Delta{R^0}}=0$
   \STATE Send initial model $G^0$ to each client
   \FOR{$each\ round\ t\gets 1,2,...,T$}
     \STATE $S_t\gets random\ select\ M\ clients$
     \FOR{$each\ client\ i\ \in  S_t\ in\ parallel$}
        \STATE $\Delta{R^t_i} \gets ClientUpdate(i,\quad t, \quad  \overline{\Delta{R^{t-1}}})$
      \ENDFOR
      \STATE RPN aggregation:  $\overline{\Delta{R^t}} = \frac{1}{M}\sum_{i=1}^{M}{\Delta{R^t_i}}$  
   \ENDFOR
\end{algorithmic}
\end{algorithm}

The main function of federated server is used to aggregate the model parameters uploaded from all selective clients, when aggregation completed, the new updated global model parameters are then subsequently sent back to the clients for next iteration. In this section, we analyze our RPN algorithm from the perspective of federated server, in our new strategy, we don't aggregate raw model parameters, but residual pooling network parameters instead, since only compressed parameters are transmitted, the amount of data downloaded will be significantly reduced. 

First, we initialize global model as $G^0$, and send it to all clients, also set $\overline{\Delta{R^{0}}}=0$. For each round $t$ at server side, also for each client $i$, we send a triple ($i$, $t$ and $\overline{\Delta{R^{t-1}}}$) to all selective clients for local model training, wait until we get response $\Delta{R^t_i}$ from all selective client, the global aggregation can be expressed as follows:

\begin{equation}\label{rpn_aggregation}
    \overline{\Delta{R^t}} = \frac{1}{M}\sum_{i=1}^{M}{\Delta{R^t_i}}
\end{equation}

Repeated this cycle until model converges or maximum number of iterations are satisfied. The entire federated server algorithm can refer to Algorithm \ref{algo_server}.

\subsection{Residual Pooling Network - Client perspective}
Typical federated learning requires each client updates the local model based on local dataset, and then send model parameters to server side for aggregation, however, as previously discussed, not all parameters are sensitive to model updates in distributed model training. In this section, We will analyze RPN algorithm from the perspective of the client, and show how the amount of data uploaded will be significantly reduced. 

Without loss of generality, suppose we start at round $t$. The first step is to recover local model $G_i^{t-1}$, based on $G_i^0$ and the sum of $\overline{\Delta{R^j}}$, where $j$ from 1 to $t-1$, which means:

\begin{equation}\label{recover}
    G^{t-1}_i=G^{0}_i + \sum_{j=1}^{t-1}{\overline{\Delta{R^{j}}}}
\end{equation}

After that, based on local datasets $D_i$, we do local model update, change the model from $G^{t-1}_i$ to $G^{t}_i$, and calculate residual network $R^{t}_i =  G^{t}_i -  G^{t-1}_i$, compress $R^{t}_i$ using spatial mean pooling technique, get the final RPN model $\Delta{R^{t}_i}$, send it back to server side for aggregation. The entire federated client algorithm can refer to Algorithm \ref{algo_client}.

\begin{algorithm}
   \caption{RPN: FL\_Client (ClientUpdate)}
   \label{algo_client}
   \setstretch{1.3} 
\begin{algorithmic}
   \STATE {\bfseries Input:} client id $i$;\ round $t$;\ $\overline{\Delta{R^{t-1}}}$
   \STATE {\bfseries Output:} model parameters that sent back to server
   \STATE Recover model: $G^{t-1}_i=G^{0}_i + \sum_{j=1}^{t-1}{\overline{\Delta{R^{j}}}}$
   \FOR{$each\ local\ epoch\ e\gets 1,2...E$}
     \STATE $X \gets random\ sample\ dataset\ with\ size\ B$ 
     \STATE $w^{t}_i \xleftarrow{} w^{t-1}_i - \eta * \nabla{l(y_i, f(D_i;{w}))}$ 
   \ENDFOR 
    \STATE Let $G^{t}_i=f(w^{t}_i)$ 
    \STATE Calculate residual model $R^{t}_i=G^{t}_i - G^{t-1}_i$ 
    \STATE Compress $R^{t}_i$ with spatial pooling, and get $\Delta{R^t_i}$ 
    \STATE Send $\Delta{R^t_i}$ back to server side
   
\end{algorithmic}
\end{algorithm}

\begin{table*}
\caption{Mnist data distribution of each federated client}\label{tab:mnist_dist}
\centering
\begin{tabular}{|c|c|c|c|c|c|c|c|c|c|c|c|}
\hline
    &  & \multicolumn{10}{|c|}{label distribution}   \\ \cline{3-12}
client   & total image number & 0 & 1 & 2 & 3 & 4 & 5 & 6 & 7 & 8 & 9  \\ 
\hline
client 1    & 5832 & 763 & 586 & 423 & 533 & 582 & 607 & 690 & 568 & 539 & 541  \\
\hline
client 2    & 5230 & 649 & 798 & 518 & 330 & 429 & 380 & 320 & 731 & 626 & 449  \\
\hline
client 3    & 6190 & 363 & 724 & 722 & 421 & 611 & 612 & 531 & 669 & 708 & 829  \\
\hline
client 4    & 6832 & 518 & 831 & 605 & 1082 & 637 & 607 & 996 & 586 & 701 & 269  \\
\hline
client 5    & 5903 & 420 & 570 & 588 & 763 & 288 & 688 & 611 & 641 & 600 & 734  \\
\hline
client 6    & 5598 & 518 & 763 & 563 & 597 & 591 & 531 & 328 & 608 & 421 & 678  \\
\hline
client 7    & 6239 & 611 & 524 & 855 & 628 & 789 & 664 & 511 & 709 & 790 & 158  \\
\hline
client 8    & 6166 & 417 & 449 & 511 & 666 & 737 & 508 & 1080 & 563 & 563 & 672  \\
\hline
client 9    & 5298 & 523 & 960 & 739 & 707 & 432 & 498 & 285 & 420 & 488 & 246  \\
\hline
client 10    & 6712 & 1141 & 537 & 434 & 404 & 746 & 326 & 566 & 770 & 425 & 1373 \\ 
\hline
\end{tabular}
\label{tab:mnist}
\end{table*}

\subsection{Algorithm Correctness Analysis}

\textbf{\textit{Lemma 1.}} Denote $w^{t-1}$ is global model parameters after $(t-1)$ times aggregation, $w^{t-1}_i$ is the local model parameters of client $i$ before training on round $t$, and $w^{t}_i$ is the corresponding parameters after training on round $t$, under \textbf{iid} condition, we should have:

\begin{equation}\label{iid}
    \mathbb{E}[\frac{1}{M}\sum_{i=1}^{M}{({w^t_i} - w^{t-1}_i)}] = \mathbb{E}[\frac{1}{M}\sum_{i=1}^{M}{\Delta{R^{t}_i}}]
\end{equation}

In order to validate the correctness of previous equation, we can notice that for each client $C_i$ and local model $G^t_i$, mean operation on residual model parameters $(R^t_i={w^t_i} - w^{t-1}_i)$ does not affect expectation, this implies the following equation is satisfied:

\begin{equation}\label{iid2}
    \mathbb{E}[{({w^t_i} - w^{t-1}_i)}] = \mathbb{E}[{\Delta{R^{t}_i}}]
\end{equation}

According to equation \ref{iid2}, it is easy to prove equation \ref{iid} is correct.

\bigskip
\textbf{\textit{Lemma 2.}} Suppose we have $N$ clients in our federated learning cluster, Algorithm \ref{algo_server} and Algorithm \ref{algo_client} can guarantee to recover local model on each round.

To reduce the amount of data being download, the server send compressed model $\overline{\Delta{R^t}}$ to client (see algorithm \ref{algo_server}), whenever the client receive $\overline{\Delta{R^t}}$, we can not apply it directly, because we have to recover to local model $G^t_i$, the first step in algorithm \ref{algo_client} show how to perform this process, the correctness can be guaranteed by the following equation:

\begin{equation}
\begin{aligned}
\label{iid3}
     G^0 + \sum_{j=1}^{n}{\overline{R^j}} &= G^0 + \sum_{j=1}^{k}{\overline{R^j}} - \sum_{j=1}^{k}{\overline{R^j}} + \sum_{j=1}^{n}{\overline{R^j}} \\ &= ( G^0 + \sum_{j=1}^{k}{\overline{R^j}}) + (\sum_{j=1}^{n}{\overline{R^j}} - \sum_{j=1}^{k}{\overline{R^j}}) \\ &= G^k +  \sum_{j=k+1}^{n}{\overline{R^j}} = G^n
\end{aligned}    
\end{equation}

Since $k$ and $n$ are random, which satisfy $k \leq n$, using mathematical induction, we can easy prove the correctness.

\subsection{Performance Analysis}
The overall communication costs consist of two aspects: the upload from client to server and download from server to client.

\begin{itemize}
\item \textbf{Upload:} According to Algorithm \ref{algo_client}, the total number of uploaded data is equal to: 

\begin{equation}\label{upload_amount}
   S_{upload} = \sum_{j=1}^{M}{|\Delta{R^t_j}|}
\end{equation}

Since we execute spatial mean pooling on convolutional layer, suppose the kernel shape of model $G_i^t$ is $(n, s, s, m)$, where $n$ is the number of filter input, $m$ is the number of filter output size, $s$ is kernel size. after compressed, the kernel shape of model $\Delta{R^t_i}$ is $(n, 1, 1, m)$. Obviously, compared with standard federated learning, the data upload amount reduce $(s*s)$ times for each convolutional kernel.  

\item \textbf{Download:} According to Algorithm \ref{algo_server}, the total number of downloaded data is equal to:

\begin{equation}\label{download_amount}
   S_{download} = |\overline{\Delta{R^{t}}}|
\end{equation}

$\overline{\Delta{R^{t}}}$ is the aggregation of all RPN model $\Delta{R^t_i}$
collected from selective clients, the shape of model $\overline{\Delta{R^{t}}}$ is equal to $\Delta{R^t_i}$, as previously discussed, the data download amount also reduce $(s*s)$ times for each convolutional kernel.  

\end{itemize}

\section{Experiments}

We evaluate our approach on three different tasks: image classification, object detection and semantic segmentation, we will compare the performance between standard federated learning and RPN. All our experiments are run on a federated learning cluster consists of one server side and 10 clients, each of which are equipped with 8 NVIDIA TeslaV100 GPUs.

\subsection{Experiments on MNIST classification}

We evaluate classification performance on mnist dataset using a 5-layers CNN model, which contains total 1,199,882 parameters, after using RPN technique, the transmitted data can drop to 1,183,242, since most of the parameters are came from full connection layer, the compression effect is not obvious enough. we split the data into 10 parts randomly, and send it to each client for training datasets, the data distribution of each client is shown in table \ref{tab:mnist}. 

\begin{figure}[htb]
\centering

\includegraphics[width=8.2cm]{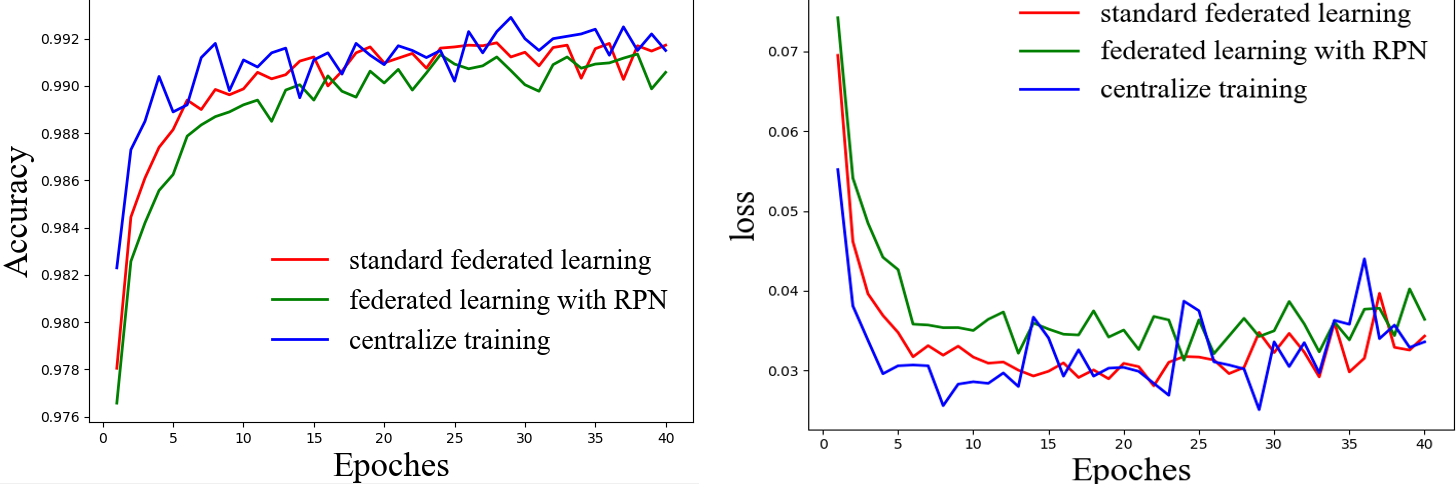}

\caption{Experiments on MNIST dataset}
\label{fig:map_mnsit}
\end{figure}

The results are shown in Figure \ref{fig:map_mnsit}, as we can see, RPN can achieve almost the same effect as standard federated learning.

Our experiment performance summarize in table \ref{mnist_performance}:

\begin{table}
\caption{experiments performance on mnist classification}
\centering                                                                                                
\begin{tabular}{|c|c|c|}
\hline
 & \multicolumn{2}{|c|}{MNIST} \\ \cline{2-3}

 Model &  Parameters transmission & Accuracy  \\ 
 \hline
FL  & 1,199,882 & 0.991  \\ 
\hline
RPN & 1,183,242 & 0.9912 \\
\hline\hline
Performance & $\downarrow\textbf{1.4\%}$ & --- \\ 
\hline
\end{tabular}
\label{mnist_performance}
\end{table}

\begin{table*}
\caption{PASCAL VOC data distribution of each federated client}\label{tab:PASCAL_VOC}
\centering
\begin{tabular}{|c|c|c|c|c|c|c|c|c|c|c|}
\hline
    &  \multicolumn{10}{|c|}{client id}   \\ \cline{2-11}
label   & client 1 & client 2 & client 3 & client 4 & client 5 & client 6 & client 7 & client 8 & client 9 & client 10  \\ 
\hline
Aeroplane & 53 & 32 & 71 & 66 & 58 & 61 & 38 & 45 & 39 & 60 \\
\hline 
Bicycle & 21 & 62 & 33 & 48 & 39 & 56 & 57 & 68 & 42 & 40 \\
\hline 
Bird & 61 & 68 & 103 & 52 & 50 & 75 & 37 & 43 & 59 & 50 \\
\hline 
Boat & 31 & 42 & 71 & 52 & 23 & 36 & 43 & 54 & 29 & 40 \\
\hline 
Bottle & 71 & 82 & 38 & 54 & 65 & 98 & 37 & 77 & 61 & 81 \\
\hline 
Bus & 31 & 12 & 53 & 64 & 55 & 46 & 27 & 18 & 59 & 60 \\
\hline 
Car & 123 & 108 & 45 & 64 & 72 & 151 & 77 & 48 & 89 & 103 \\
\hline 
Cat & 75 & 89 & 101 & 161 & 77 & 87 & 89 & 58 & 113 & 94 \\
\hline 
Chair & 170 & 65 & 79 & 91 & 106 & 118 & 68 & 91 & 117 & 67 \\
\hline 
Cow & 12 & 25 & 39 & 33 & 41 & 29 & 17 & 18 & 29 & 20 \\
\hline 
Diningtable & 34 & 28 & 45 & 72 & 52 & 61 & 37 & 28 & 49 & 30 \\
\hline 
Dog & 208 & 102 & 113 & 178 & 58 & 69 & 91 & 82 & 119 & 98 \\
\hline 
Horse & 43 & 25 & 31 & 49 & 72 & 28 & 19 & 31 & 27 & 42 \\
\hline 
Motorbike & 16 & 72 & 21 & 40 & 53 & 39 & 72 & 68 & 49 & 52 \\
\hline 
Person & 340 & 689 & 213 & 480 & 111 & 239 & 128 & 196 & 295 & 303 \\
\hline 
PottedPlant & 45 & 24 & 65 & 83 & 28 & 31 & 39 & 18 & 45 & 51 \\
\hline 
Sheep & 28 & 12 & 61 & 22 & 14 & 20 & 30 & 18 & 39 & 27 \\
\hline 
Sofa & 42 & 39 & 29 & 37 & 35 & 20 & 51 & 54 & 32 & 30 \\
\hline 
Train & 34 & 26 & 66 & 14 & 25 & 32 & 38 & 40 & 19 & 41 \\
\hline 
Tvmonitor & 68 & 43 & 23 & 45 & 39 & 29 & 17 & 29 & 31 & 40 \\
\hline 

\end{tabular}
\label{tab:pascal}
\end{table*}

\subsection{Experiments on Pascal-VOC object detection}

In this section, we conduct object detection task on PASCAL VOC dataset. The PASCAL Visual Object Classification (PASCAL VOC) dataset is a well-known dataset for object detection, classification, segmentation of objects and so on. There are around 10,000 images for training and validation containing bounding boxes with objects. Although, the PASCAL VOC dataset contains only 20 categories, it is still considered as a reference dataset in the object detection problem.

We conduct object detection task on pascal voc dataset with YOLOV3. YOLOV3 contrains 107 layer with 75 are convolutional layers. same as mnist experiment, we split the data into 10 parts randomly, and send it back to each client for training data. The data distribution of each client is shown in table \ref{tab:PASCAL_VOC}. 

\begin{figure}[htb]
\centering

\includegraphics[width=8.2cm]{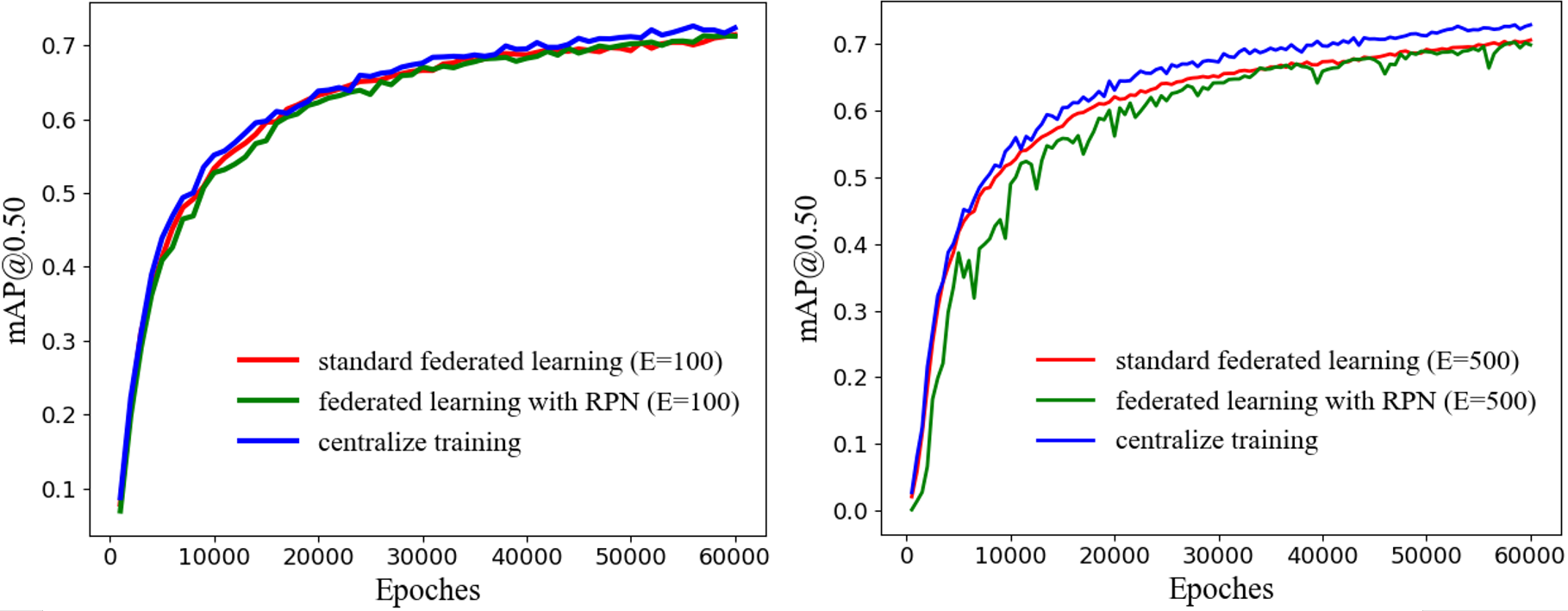}

\caption{Experiments on PASCAL VOC object detection}
\label{fig:map_pascal}
\end{figure}

Our experiment results are shown in Figure \ref{fig:map_pascal}. We use mean average precision (mAP) as our performance metrics, mAP is a popular metric in measuring the accuracy of object detectors like Faster R-CNN, SSD, etc. As we can see, rpn will shuffe fluctuation in the early stage, but as the iterations continue, the performance becomes stable, and shows shallow performance gap between rpn and standard federated learning.   

Our experiment performance summarize in table \ref{pas_performance}, since in YOLOV3 network architecture, about 70\% layers are convolutional layer, compared with mnist classification, our RPN compression strtegy is more obvious, we can reduce model parameters from 61,895,776 to 12,382,560, while only one percent of performance is declining.

\begin{table}
\caption{experiments performance on PASCAL VOC Object detection}
\centering                                                                                                
\begin{tabular}{|c|c|c|}
\hline
 & \multicolumn{2}{|c|}{PASCAL VOC} \\ \cline{2-3}

 Model &  Parameters transmission & Accuracy  \\ 
 \hline
FL  & 61,895,776 & 0.7077  \\ 
\hline
RPN  &  12,382,560 & 0.7002\\
\hline\hline
Performance  & $\downarrow\textbf{80\%}$ & $\downarrow\textbf{1\%}$ \\ 
  \hline
\end{tabular}
\label{pas_performance}
\end{table}

\subsection{Experiments on Pascal-VOC Semantic segmentation}

In this section, we conduct Semantic segmentation task on PASCAL VOC dataset, and use fully convolutional networks (FCN) as our model. FCN can use different model architecture for feature extraction, such as ResNet, vggnet, DenseNet, etc. In our experiment, we use Resnet50 as our based models. The data distribution of each client is shown in table \ref{tab:PASCAL_VOC}. 

\begin{figure}[htb]
\centering

\includegraphics[width=8.2cm]{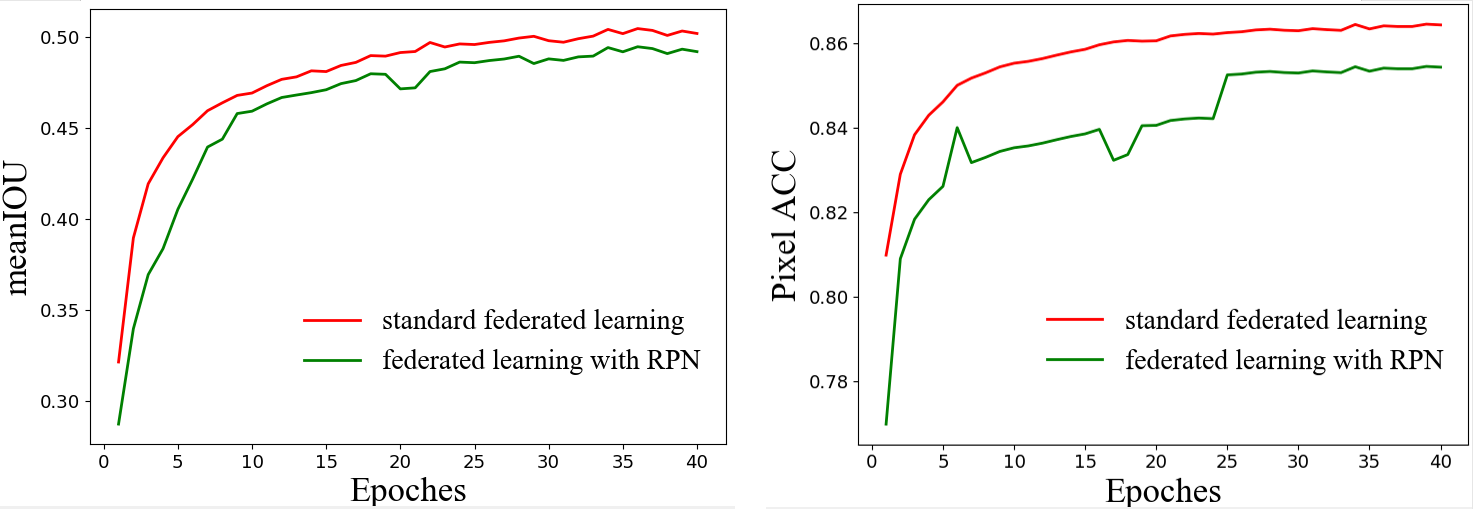}

\caption{Experiments on PASCAL VOC semantic segmentation}
\label{fig:seg}
\end{figure}

Our experiment results are shown in Figure \ref{fig:seg}. We use meanIOU and pixel accuracy as our performance metrics.

Our experiment performance summarize in table \ref{seg_performance}, we can reduce model parameters from 23,577,621 to 13,325,281, while only 2.6 percent of performance is declining for meanIOU metric, and 1.16 percent of performance is declining for pixel accuracy.

\begin{table}
\caption{experiments performance on PASCAL VOC Object detection}
\centering                                                                                                
\begin{tabular}{|c|c|c|c|}
\hline
 & \multicolumn{3}{|c|}{PASCAL VOC} \\ \cline{2-4}

 Model &  Parameters transmission & meanIOU & Pixel ACC  \\ 
 \hline
FL  & 23,577,621 & 0.5077 & 0.864 \\ 
\hline
RPN  &  13,325,281 & 0.4942 & 0.854 \\
\hline\hline
Performance  & $\downarrow\textbf{44\%}$ & $\downarrow\textbf{2.6\%}$ & $\downarrow\textbf{1.16\%}$ \\ 
  \hline
\end{tabular}
\label{seg_performance}
\end{table}

\section{CONCLUSION and Future Work}
In this paper, we propose a new compression strategy to improve communication efficiency of federated learning. we test our approach on three different tasks, including classification, object detection and semantic segmentation, the experiments show that RPN not only reduce data transmission effectively, but also achieve almost the same performance as compared to standard federated learning. Most importantly, RPN is n end-to-end procedure, which makes it easy to deploy in real-world application without human intervention. In the future, we will combine other compression strategies to improve communication efficiency.

\bibliography{ecai}

\begin{thebibliography}{10}

\bibitem{aji2017sparse}
Alham~Fikri Aji and Kenneth Heafield, `Sparse communication for distributed
  gradient descent', {\em arXiv preprint arXiv:1704.05021}, (2017).

\bibitem{DBLP:journals/corr/abs-1802-04434}
Jeremy Bernstein, Yu{-}Xiang Wang, Kamyar Azizzadenesheli, and Anima
  Anandkumar, `signsgd: compressed optimisation for non-convex problems', {\em
  CoRR}, {\bf abs/1802.04434}, (2018).

\bibitem{DBLP:journals/corr/abs-1810-05291}
Jeremy Bernstein, Jiawei Zhao, Kamyar Azizzadenesheli, and Anima Anandkumar,
  `signsgd with majority vote is communication efficient and byzantine fault
  tolerant', {\em CoRR}, {\bf abs/1810.05291}, (2018).

\bibitem{DBLP:journals/corr/BojanowskiGJM16}
Piotr Bojanowski, Edouard Grave, Armand Joulin, and Tomas Mikolov, `Enriching
  word vectors with subword information', {\em CoRR}, {\bf abs/1607.04606},
  (2016).

\bibitem{DBLP:journals/corr/abs-1902-01046}
Keith Bonawitz, Hubert Eichner, Wolfgang Grieskamp, Dzmitry Huba, Alex
  Ingerman, Vladimir Ivanov, Chlo{\'{e}} Kiddon, Jakub Konecn{\'{y}}, Stefano
  Mazzocchi, H.~Brendan McMahan, Timon~Van Overveldt, David Petrou, Daniel
  Ramage, and Jason Roselander, `Towards federated learning at scale: System
  design', {\em CoRR}, {\bf abs/1902.01046}, (2019).

\bibitem{Bonawitz:2017:PSA:3133956.3133982}
Keith Bonawitz, Vladimir Ivanov, Ben Kreuter, Antonio Marcedone, H.~Brendan
  McMahan, Sarvar Patel, Daniel Ramage, Aaron Segal, and Karn Seth, `Practical
  secure aggregation for privacy-preserving machine learning', in {\em
  Proceedings of the 2017 ACM SIGSAC Conference on Computer and Communications
  Security}, CCS '17, pp. 1175--1191, New York, NY, USA, (2017). ACM.

\bibitem{DBLP:journals/corr/abs-1812-07210}
Sebastian Caldas, Jakub Konecn{\'{y}}, H.~Brendan McMahan, and Ameet Talwalkar,
  `Expanding the reach of federated learning by reducing client resource
  requirements', {\em CoRR}, {\bf abs/1812.07210}, (2018).

\bibitem{DBLP:journals/corr/ChenSX17}
Yudong Chen, Lili Su, and Jiaming Xu, `Distributed statistical machine learning
  in adversarial settings: Byzantine gradient descent', {\em CoRR}, {\bf
  abs/1705.05491}, (2017).

\bibitem{DBLP:journals/corr/abs-1810-04805}
Jacob Devlin, Ming{-}Wei Chang, Kenton Lee, and Kristina Toutanova, `{BERT:}
  pre-training of deep bidirectional transformers for language understanding',
  {\em CoRR}, {\bf abs/1810.04805}, (2018).

\bibitem{Goodfellow:2016:DL:3086952}
Ian Goodfellow, Yoshua Bengio, and Aaron Courville, {\em Deep Learning}, The
  MIT Press, 2016.

\bibitem{DBLP:journals/corr/HanMD15}
Song Han, Huizi Mao, and William~J. Dally, `Deep compression: Compressing deep
  neural network with pruning, trained quantization and huffman coding', in
  {\em 4th International Conference on Learning Representations, {ICLR} 2016,
  San Juan, Puerto Rico, May 2-4, 2016, Conference Track Proceedings}, (2016).

\bibitem{DBLP:journals/corr/HannunCCCDEPSSCN14}
Awni~Y. Hannun, Carl Case, Jared Casper, Bryan Catanzaro, Greg Diamos, Erich
  Elsen, Ryan Prenger, Sanjeev Satheesh, Shubho Sengupta, Adam Coates, and
  Andrew~Y. Ng, `Deep speech: Scaling up end-to-end speech recognition', {\em
  CoRR}, {\bf abs/1412.5567}, (2014).

\bibitem{DBLP:journals/corr/HeZR014}
Kaiming He, Xiangyu Zhang, Shaoqing Ren, and Jian Sun, `Spatial pyramid pooling
  in deep convolutional networks for visual recognition', {\em CoRR}, {\bf
  abs/1406.4729}, (2014).

\bibitem{DBLP:journals/corr/HeZRS15}
Kaiming He, Xiangyu Zhang, Shaoqing Ren, and Jian Sun, `Deep residual learning
  for image recognition', {\em CoRR}, {\bf abs/1512.03385}, (2015).

\bibitem{DBLP:journals/corr/IoffeS15}
Sergey Ioffe and Christian Szegedy, `Batch normalization: Accelerating deep
  network training by reducing internal covariate shift', {\em CoRR}, {\bf
  abs/1502.03167}, (2015).

\bibitem{DBLP:journals/corr/KonecnyMYRSB16}
Jakub Konecn{\'{y}}, H.~Brendan McMahan, Felix~X. Yu, Peter Richt{\'{a}}rik,
  Ananda~Theertha Suresh, and Dave Bacon, `Federated learning: Strategies for
  improving communication efficiency', {\em CoRR}, {\bf abs/1610.05492},
  (2016).

\bibitem{Krizhevsky:2012:ICD:2999134.2999257}
Alex Krizhevsky, Ilya Sutskever, and Geoffrey~E. Hinton, `Imagenet
  classification with deep convolutional neural networks', in {\em Proceedings
  of the 25th International Conference on Neural Information Processing Systems
  - Volume 1}, NIPS'12, pp. 1097--1105, USA, (2012). Curran Associates Inc.

\bibitem{0483bd9444a348c8b59d54a190839ec9}
Yann LeCun, Yoshua Bengio, and Geoffrey Hinton, `Deep learning', {\em Nature},
  {\bf 521}(7553),  436--444, (5 2015).

\bibitem{lim2019federated}
Wei Yang~Bryan Lim, Nguyen~Cong Luong, Dinh~Thai Hoang, Yutao Jiao, Ying-Chang
  Liang, Qiang Yang, Dusit Niyato, and Chunyan Miao.
\newblock Federated learning in mobile edge networks: A comprehensive survey,
  2019.

\bibitem{DBLP:journals/corr/McMahanMRA16}
H.~Brendan McMahan, Eider Moore, Daniel Ramage, and Blaise~Ag{\"{u}}era
  y~Arcas, `Federated learning of deep networks using model averaging', {\em
  CoRR}, {\bf abs/1602.05629}, (2016).

\bibitem{mikolov2013efficient}
Tomas Mikolov, Kai Chen, Greg Corrado, and Jeffrey Dean.
\newblock Efficient estimation of word representations in vector space, 2013.
\newblock cite arxiv:1301.3781.

\bibitem{DBLP:journals/corr/abs-1903-02891}
Felix Sattler, Simon Wiedemann, Klaus{-}Robert M{\"{u}}ller, and Wojciech
  Samek, `Robust and communication-efficient federated learning from non-iid
  data', {\em CoRR}, {\bf abs/1903.02891}, (2019).

\bibitem{strom2015scalable}
Nikko Strom, `Scalable distributed dnn training using commodity gpu cloud
  computing', in {\em Sixteenth Annual Conference of the International Speech
  Communication Association}, (2015).

\bibitem{DBLP:journals/corr/SureshYMK16}
Ananda~Theertha Suresh, Felix~X. Yu, H.~Brendan McMahan, and Sanjiv Kumar,
  `Distributed mean estimation with limited communication', {\em CoRR}, {\bf
  abs/1611.00429}, (2016).

\bibitem{216799}
Zeyi Tao and Qun Li, `esgd: Communication efficient distributed deep learning
  on the edge', in {\em {USENIX} Workshop on Hot Topics in Edge Computing
  (HotEdge 18)}, Boston, MA, (July 2018). {USENIX} Association.

\bibitem{tsuzuku2018variance}
Yusuke Tsuzuku, Hiroto Imachi, and Takuya Akiba, `Variance-based gradient
  compression for efficient distributed deep learning', {\em arXiv preprint
  arXiv:1802.06058}, (2018).

\bibitem{DBLP:journals/corr/OordDZSVGKSK16}
A{\"{a}}ron van~den Oord, Sander Dieleman, Heiga Zen, Karen Simonyan, Oriol
  Vinyals, Alex Graves, Nal Kalchbrenner, Andrew~W. Senior, and Koray
  Kavukcuoglu, `Wavenet: {A} generative model for raw audio', {\em CoRR}, {\bf
  abs/1609.03499}, (2016).

\bibitem{Voigt:2017:EGD:3152676}
Paul Voigt and Axel von~dem Bussche, {\em The EU General Data Protection
  Regulation (GDPR): A Practical Guide}, Springer Publishing Company,
  Incorporated, 1st edn., 2017.

\bibitem{wang2018atomo}
Hongyi Wang, Scott Sievert, Zachary Charles, Shengchao Liu, Stephen Wright, and
  Dimitris Papailiopoulos.
\newblock Atomo: Communication-efficient learning via atomic sparsification,
  2018.

\bibitem{CMFL2019}
Luping Wang, Wei Wang, and Bo~Li, `Cmfl: Mitigating communication overhead for
  federated learning', in {\em 2019 IEEE 39th International Conference on
  Distributed Computing Systems}, Dallas, TX, (2019).

\bibitem{DBLP:journals/corr/abs-1902-04885}
Qiang Yang, Yang Liu, Tianjian Chen, and Yongxin Tong, `Federated machine
  learning: Concept and applications', {\em CoRR}, {\bf abs/1902.04885},
  (2019).

\bibitem{DBLP:journals/corr/ZhangCJ16}
Yu~Zhang, William Chan, and Navdeep Jaitly, `Very deep convolutional networks
  for end-to-end speech recognition', {\em CoRR}, {\bf abs/1610.03022}, (2016).

\end{thebibliography}
\end{document}